\documentclass[letterpaper, 10 pt, conference]{ieeeconf}

\IEEEoverridecommandlockouts
\makeatletter\let\NAT@parse\undefined\makeatother
\usepackage[numbers,sort&compress]{natbib}

\usepackage{amsmath,amssymb,amsfonts}
\usepackage{algorithmic}
\usepackage{graphicx}
\usepackage{textcomp}
\usepackage{algorithm}
\usepackage{float}
\usepackage{booktabs,multirow}
\usepackage[dvipsnames]{xcolor}
\usepackage{url}
\usepackage{tabularx}
\usepackage{siunitx}
\usepackage{tikz}
\usetikzlibrary{arrows.meta}

\usepackage{microtype}
\usepackage{caption}

\usepackage[colorlinks=true,citecolor=cyan,linkcolor=black,urlcolor=black,bookmarks=false,pdfusetitle]{hyperref}

\usepackage[capitalise]{cleveref}

\newcommand{\rrtdiff}[1]{\textcolor{blue}{#1}}

\begin{document}

\title{\LARGE \bf
PoliMop: Policy-in-the-Loop Motion Planning in 3D Spaces for Mobile Robots
}
\author{%
Yael Fassbind, Valentin N. Hartmann, Huang Su, Yijiang Huang, Stelian Coros%
\thanks{All authors are with the Computational Robotics Lab,
ETH Zurich, Switzerland.}%
}

\maketitle

\begin{abstract}
Reaching a goal state in a complex environment often requires a robot to overcome challenging terrain and obstacles.
Which routes are accessible depends on the robot’s physical design, the control policy it uses, and how it modifies its surroundings.
We present a kinodynamic rapidly-exploring random tree (RRT) planner that uses physics simulation to exploit skills encoded in the control policy, which allows it to find a path in the scene with respect to the robot's capabilities.
The planner samples states from a reduced state space in 3D around the scene and expands the RRT toward each sample using policy rollouts.
We guide sampling with a preplanned path for a simple mass-point model to focus exploration on promising regions leading toward the goal state, reducing planning time by up to $\textbf{63\%}$.

Simulation experiments with quadrupeds, humanoids, and wheeled robots demonstrate that our approach supports drop-in integration of various robots and control policies, exploiting their physical and policy capabilities without modifying the underlying planning algorithm.
Predefined environmental modifications, such as placing a plank across a gap to build a bridge, will automatically be picked up by the planner as the robot's reachable space expands.
Videos and code are available \href{https://ariael21.github.io/polimop.html}{\textcolor{blue}{here}}.
\end{abstract}

\section{Introduction}
\label{sec:introduction}

Motion planning is typically treated as the problem of finding a feasible path through a fixed environment.
For robots operating in complex terrain, however, reachability depends not only on the scene geometry but also on the robot's body and brain, i.e., its mechanical design and control policies.
A path that is feasible for a quadruped equipped with a parkour control policy (e.g.,~\cite{hoeller2024parkour}) may be infeasible for a wheeled robot or for the same quadruped using a simpler locomotion policy.
Moreover, a robot can intentionally change the scene by manipulating objects to actively extend its reachable space.
For example, a robot may place a plank across a gap to create a bridge to make a final platform reachable.
Thus, reachability is not a fixed property of the scene: it emerges from the combination of the robot, the selected policy, and the current environment state.
Ultimately, our vision is to enable robots not only to exploit the capabilities of their control policies to traverse challenging terrain but also to {\it deliberately modify their environment} to reach a goal or complete a task.
In this work, we take a step toward this goal by asking: how can a motion planner find paths in environments whose reachable regions change through prescribed manipulation actions and policy selections?

\begin{figure}[t]
    \centering
    \includegraphics[width=\columnwidth]{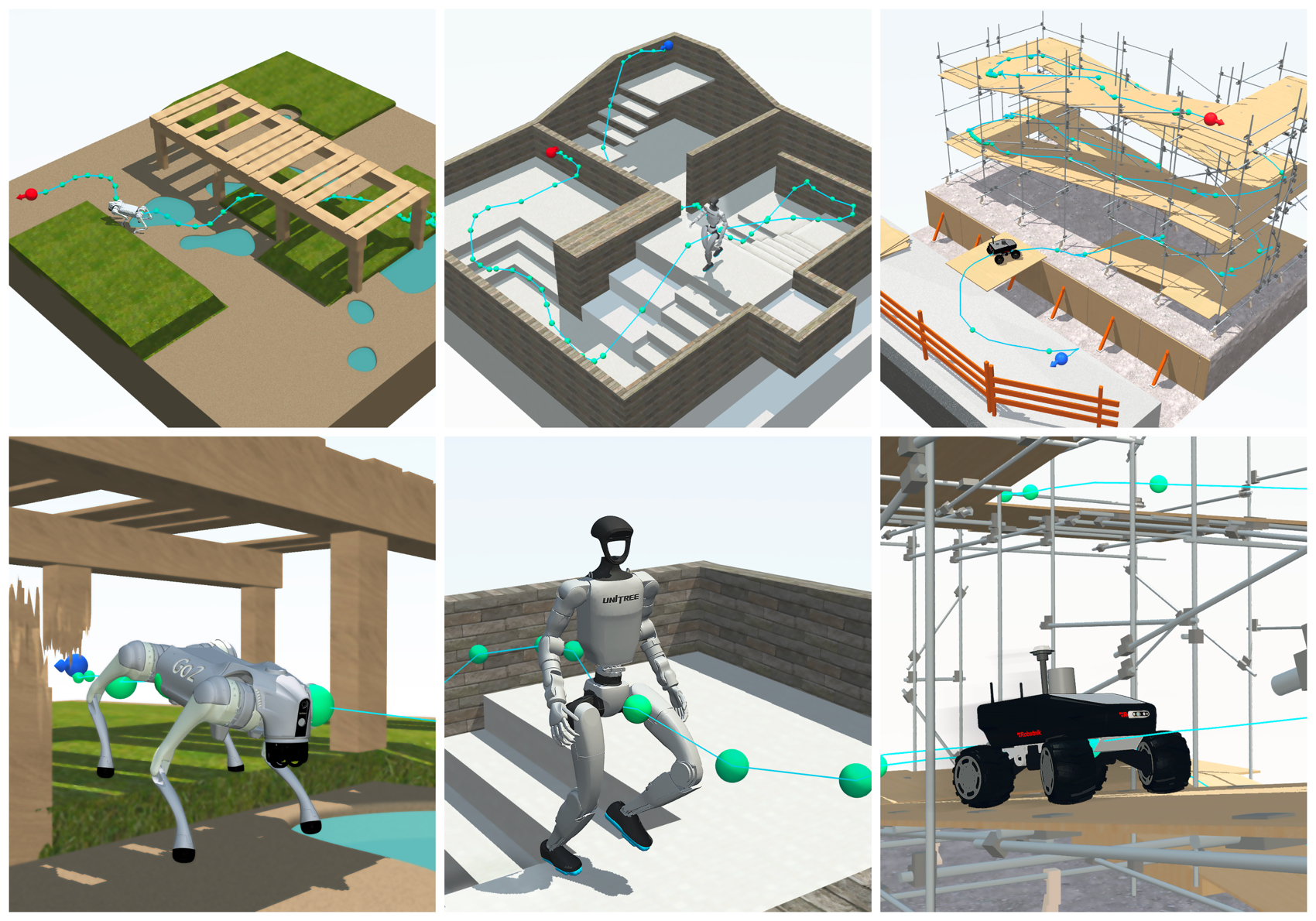}
    \caption{Our policy-in-the-loop motion planner enables diverse robots to traverse challenging environments. A quadruped (left), humanoid (center), and wheeled robot (right) navigate low passages, stairs, and narrow scaffold ramps.}
    \label{fig:teaser}
\end{figure}

Many conventional motion-planning approaches use explicit representations of traversable terrain, such as 2.5D elevation maps combined with robot-specific traversability estimates~\cite{wermelinger2016navigation,wellhausen2023artplanner}.
A single-layer elevation map stores one terrain height per horizontal location and therefore cannot perform motion planning on vertically overlapping surfaces (\cref{fig:teaser}, right).
Additionally, the planner should automatically exploit the robot's capabilities encoded in its control policy.
To achieve both, our planner samples target positions throughout the 3D scene and uses a physics simulator as a state-transition mechanism to roll out motions generated by the control policy.
This enables us to plan through multi-level structures, as illustrated in \cref{fig:teaser}, while accounting for the selected robot--policy pair's capabilities without the requirement of an explicit parameterization of the reachable space that the planner needs to sample from.

We formulate our approach in the framework of kinodynamic RRT \cite{lavalle1998rrt,lavalle2001kinodynamic} as its incremental tree structure provides a natural way to explore an environment through policy-driven rollouts.
During each expansion, we use a policy rollout to produce a motion such that the robot moves toward a sampled target in 3D.
Therefore, the RRT explores the reachable space produced by the selected robot--policy pair, allowing capabilities such as walking up stairs (\cref{fig:teaser}) or even climbing walls to emerge without being modeled explicitly.
This also means that our planner can adapt to new robot models, control policies, or 3D scene changes in a plug-and-play manner, without needing to change the motion-planning framework.

Our contributions include:
\begin{itemize}
    \item A \textbf{policy-in-the-loop kinodynamic Rapidly-exploring Random Tree (RRT)} that uses a physics simulator to evaluate control-policy rollouts, such that exploration reflects the capabilities of the selected robot--policy pair in the current environment.
    \item A \textbf{path-guided sampling strategy} that reduces planning time by focusing on promising regions of the state space using a simplified mass-point model.
\end{itemize}
We evaluate our planner using a library of benchmark scenes covering stairs, labyrinths, slopes, narrow passages, and environmental changes under predefined task plans.
Using this benchmark, we demonstrate that our planner finds solutions reliably with reasonable computation times, exploits robot--policy capabilities, and adapts to environmental changes to reach otherwise inaccessible goals.

\section{Related Work}
\label{sec:related-work}

\paragraph*{Planning with a Robot Model}
\label{par:model-based-planning}
Legged systems display agility to cross terrain that wheeled platforms cannot, first through carefully engineered leg mechanics and hand-tuned control laws~\cite{raibert1986balance}, then through whole-body motion optimization~\cite{bledt2018cheetah3}, and most recently through policies trained by reinforcement learning in simulation~\cite{rudin2022minutes,hoeller2024parkour}.
However, a global planner needs to be used in tandem with these controllers to enable robots to autonomously determine a path in unstructured terrain.

One possibility is then formulating a trajectory optimization problem over the full kinematics and dynamics, the terrain geometry, and the contact forces that couple them, which synthesizes walking, climbing, and object manipulation from a goal specification alone~\cite{mordatch2012cio,posa2014contact}.
The resulting programs are large, non-convex, densely constrained, and hard to solve.

\paragraph*{Kinodynamic Sampling-Based Planning}
\label{par:kinodynamic}
Sampling-based planners such as Rapidly-exploring Random Trees (RRT)~\cite{lavalle1998rrt} and Probabilistic Roadmaps (PRM)~\cite{kavraki1996probabilistic}  grow a tree or graph of configurations by extending toward randomly drawn samples.
Without system differential constraints, that extension is linear interpolation; with them it requires a steering function that solves a two-point boundary value problem.
Alternatively, one can propagate sampled controls forward through the dynamics and keep wherever the system lands~\cite{li2016sst,shome2021boe}.
Recent work then mostly improves these two approaches with, e.g., extension heuristics derived for specific classes of dynamics~\cite{webb2013kinorrtstar,glassman2010lqr} or nonlinear solvers called as a subroutine inside the planner~\cite{stoneman2014embedding}, or sampling and node selection that attempt to make propagations where they are most informative~\cite{li2016sst}.
A third line sidesteps the choice altogether, expanding with precomputed motion primitives over a discretized space~\cite{sakcak2019primitives}, or joining them with a bounded discontinuity \cite{sivaramakrishnan2024rogue} that trajectory optimization later repairs~\cite{ortizharo2024idbrrt}.
Each still needs a model of the system---to steer with, to optimize over, or to generate primitives from.

\paragraph*{Simulators and Learned Policies as Propagation Primitives}
\label{par:learned-policy-as-a-state-propagation-mechanism}
In this work, we forward-propagate by rolling a control policy out in a physics simulator.
Using a physics engine as a black-box propagator and keeping the full simulator state in every node dates back to KPIECE~\cite{sucan2012treeplanner}.
Similarly, closed-loop RRT~\cite{kuwata2009clrrt} samples reference commands for a feedback controller and simulates the closed loop.

More recently, learned policies are used as the local planner of a roadmap~\cite{faust2018prmrl}, the propagation primitive of a tree~\cite{chiang2019rlrrt,sivaramakrishnan2024rogue}, or physics-verified skills composed into manipulation plans~\cite{mishani2026mosaic}.
These works use goal- or task-conditioned policies that steer on their own, replace the Euclidean metric with a learned reachability or temporal-distance estimate~\cite{chiang2019rlrrt,ichter2020belt}, and recover completeness by falling back to random controls~\cite{sivaramakrishnan2024rogue}.
We instead wrap a locomotion policy in a separate steering law, so any policy with a compatible command interface works without retraining, and read every outcome off a simulator rollout, so dynamic feasibility is neither modeled nor learned.

\paragraph*{Navigation Planning for Legged Robots}
\label{par:navigation}
Navigation planners for legged robots compile what a robot and its controller can traverse into a robot-specific intermediate representation that the search then queries: hand-tuned filters over terrain roughness, inclination, and step height~\cite{wermelinger2016navigation}, reachability abstractions paired with learned foothold scores~\cite{wellhausen2023artplanner}, or learned models of traversability, motion cost, feasibility, and forward dynamics, each fitted to one robot--controller pair from recorded experience~\cite{chavezgarcia2018traversability,roth2025fdm,luo2026feasibility}. %
Learned navigation policies instead fold a fixed catalog of skills into a high-level policy~\cite{hoeller2024parkour}.
What these share is that the capability model is built before the search and has to be rebuilt whenever the robot, the policy, or the skill set changes. %
We place the physics simulator and the deployed policy inside the search loop itself, so there is no surrogate to rebuild when the robot or the policy changes, and no fixed environment: an object the robot moves stays moved for the rest of the search.

\section{Problem Formulation}
\label{sec:problem-formulation}

In our setting, a mobile robot executes a task plan consisting of an ordered sequence of tasks.
Some tasks require motion planning, while others invoke skills that may change the state of the robot or its environment.
Since tasks are executed sequentially, the end state from one task becomes the initial state of the next.
A problem instance consists of a robot and its control policy, a task plan, and the initial state of the complete environment and robot.
We assume that the environment is fully observable and that the task plan and its corresponding success criteria are provided as input to the planner.
A task plan is specified as an ordered sequence of tasks $\mathcal{T} = \left(\tau_1,\tau_2,\ldots,\tau_K\right)$, where each task specifies its own goal constraint.
Possible tasks could be, e.g., $\operatorname{MoveTo}(z)$, $\operatorname{PickUp}(o)$, and $\operatorname{Place}(o,p)$,
where $z$ could be a pose that we need to reach, $o$ is an object to pick up, and $p$ is a position where we want to place the object.

In the following, we assume that we have access to controllers that can achieve the success conditions imposed by non-motion-planning tasks.
A solution to the problem is then an ordered sequence of task executions $\Sigma = (\sigma_1,\ldots,\sigma_K)$.
Each $\sigma_i$ is a sequence of control inputs that result in the satisfaction of the respective goal condition defined by $\tau_i$.
This abstraction permits arbitrary skills, including skills that modify the robot or the environment.
We consider a solution feasible if it can be replayed in the physics simulator from $s_0$ and satisfies every task in the prescribed order.
In this work, we are concerned with finding a feasible solution to the formulated problem, not necessarily an optimal solution.

\paragraph*{Task Completion}
\label{sec:accepted-goal-state-and-task-completion}
Motion-planning tasks with a goal state $z_{\text{goal}}$ succeed when the robot reaches a configuration with reduced state $z_{\text{state}}$ where the value of the distance function is below a threshold: $d(z_{\text{state}}, z_{\text{goal}}) < \varepsilon$. Tasks of other types define their own success conditions.

\section{Method}
\label{sec:method}

\Cref{alg:simulator-rrt} describes our approach:
Our method performs standard RRT exploration \cite{lavalle2001kinodynamic} in a reduced state space (\cref{eq:reduced-planning-state}).
Starting from the initial state, the planner repeatedly samples a reduced state $z_\mathrm{rnd}$ and selects the tree node $n_\mathrm{nearest}$ that minimizes the distance metric (\cref{eq:weighted-squared-distance}) to this sample.
The complete simulator state $s_\mathrm{nearest}$ is restored from the selected nearest node.
A high-level controller generates a command that steers the robot toward the sample.
The low-level controller, a control policy in the default case, turns this command into actuator inputs for the simulation rollout.
If the resulting robot state satisfies the validity constraints, it is added to the tree.
This process continues until a state lies within the goal threshold \(\varepsilon\).

\begin{algorithm}[t]
\caption{Dual-State Policy-in-the-Loop RRT. Blue text highlights differences to standard RRT.}
\label{alg:simulator-rrt}

\begin{algorithmic}[1]
\STATE $z_{\mathrm{start}}, z_{\mathrm{goal}} \gets$ reduced initial state and goal state
\STATE \rrtdiff{$q_G \gets$ unreached waypoints of guidance path $P_{\mathrm{ref}}$}
\STATE initialize tree $\mathcal{T}$ with the start node $n_{start}$
\STATE $n_{\mathrm{closest}}
    \gets \arg\min_{n \in \mathcal{T}}
    d\!\left(z(n),z_{\mathrm{goal}}\right)$

\WHILE{$d\!\left(z(n_{\mathrm{closest}}),z_{\mathrm{goal}}\right) > \varepsilon$}
    \STATE $z_{\mathrm{rnd}} \gets $ sample (from $q_G$, $z_{\mathrm{goal}}$ or uniformly from $\mathcal{Z}$)
    \STATE $n_{\mathrm{nearest}} \gets \arg\min_{n \in \mathcal{T}} d\!\left(z(n),z_{\mathrm{rnd}}\right)$
    \STATE \rrtdiff{restore full simulator state $s_{\mathrm{nearest}} = s(n_{\mathrm{nearest}})$}
    \STATE \rrtdiff{$a \gets$ $\pi(o(s_\mathrm{nearest}), \phi(z_{rnd}))$}
    \STATE $s_{\mathrm{next}} =
        \mathrm{Rollout}(s_{\mathrm{nearest}},a,t_{\mathrm{rollout}})$
    \IF{the resulting state of the robot is valid}
        \STATE add the resulting state $n_{\mathrm{next}}$
            as a new node to $\mathcal{T}$
        \STATE \rrtdiff{remove reached and bypassed waypoints from $q_G$}
    \ELSE
        \STATE \rrtdiff{penalize $n_{\mathrm{nearest}}$ and its ancestors in $\mathcal{T}$}
        \STATE \rrtdiff{recursively prune exhausted non-root leaves from $\mathcal{T}$}
    \ENDIF
    \STATE $n_{\mathrm{closest}}
        \gets \arg\min_{n \in \mathcal{T}}
        d\!\left(z(n),z_{\mathrm{goal}}\right)$
\ENDWHILE
\end{algorithmic}
\end{algorithm}

\subsection{Planning in a Reduced State Space}

The simulation state $s_t$ contains the entire simulation snapshot at time $t$, including the state of the robot and the environment.
Similar to task-space RRT~\cite{shkolnik2009taskspace}, we guide exploration using a reduced robot state:
\begin{equation}
    z_t =
    \left(
        p_{t},
        v_{t},
        \psi_t
    \right)
    \in \mathcal{Z},
    \label{eq:reduced-planning-state}
\end{equation} where \(p_{t}\in\mathbb{R}^3\) and \(v_{t}\in\mathbb{R}^3\) are the position and linear velocity of the robot's center of mass, respectively, and
$\psi_t \in (-\pi, \pi)$
is its heading angle.
Each node \(n_t\) in our RRT tree contains the reduced state \(z_t = z(n_t)\) as well as the corresponding full simulation state \(s_t = s(n_t)\), which is needed for further tree expansion.
The tree is initialized with the root node $n_{start}$, which contains the starting state $z_{start} = z(n_{start})$ of the current motion-planning problem.

We transition between two simulation states by forward propagating a control input via
\begin{equation}
    \begin{aligned}
      s_{t + t_{\mathrm{rollout}}} &= \operatorname{Rollout}\left(s_t, a_t, t_{\mathrm{rollout}}\right),
    \end{aligned}
    \label{eq:simulation-rollout}
\end{equation}
for a duration \(t_{\mathrm{rollout}}\).
We obtain the actuator control from a control-policy $\pi$ that maps a command \(c_t\) and a set of environment observations \(o_t\) to the actuator control inputs $a_t = \pi(o_t, c_t)$.
We elaborate on how the command $c_t = \phi(z_t)$ is computed in \cref{sec:controllers}.
The environment observations can be arbitrarily simple or complex, and could include a heightmap surrounding the robot or only contain the joint states of the robot.
A rollout is accepted only if all required constraints are satisfied at all intermediate states as well as the endpoint.
The validity conditions are that the robot remains upright and no disallowed collisions occur.
Therefore, a valid path consists of a state sequence $(s_0,s_1,\ldots,s_n)$ such that all of them satisfy the validity conditions.

\subsection{Sampling within the Reduced State Space}
\label{sec:sampling}
In each iteration of our RRT-based algorithm, a reduced state is sampled. This sampled state guides both tree expansion and nearest-neighbor selection.
To sample from the reduced state space, we define a bounding box $\mathcal{B}_p$ around the scene.
We sample the reduced state from
\begin{equation}
    \begin{aligned}
        p_{\mathrm{rnd}} &\sim \mathcal{U}(\mathcal{B}_p), \\
        v_\mathrm{rnd} &\sim \mathcal{U}\left(-v_{\max}, v_{\max}\right), \\
        \psi_{\mathrm{rnd}} &\sim \mathcal{U}\left(-\pi, \pi\right).
    \end{aligned}
\end{equation}
Any sampled configuration for which $p_{\mathrm{rnd}}$ collides with an object in the scene is rejected.
This is done using a probe sphere that approximates the collision shape of the robot.
The sphere is moved to the sampled position $p_\text{rnd}$ and checked for collisions with the scene.

\paragraph*{Sampling Strategy}
\label{par:path-guided-sampling}

\begin{figure}[t]
    \centering
    \raisebox{-0.5\height}{%
        \includegraphics[width=0.26\columnwidth]
            {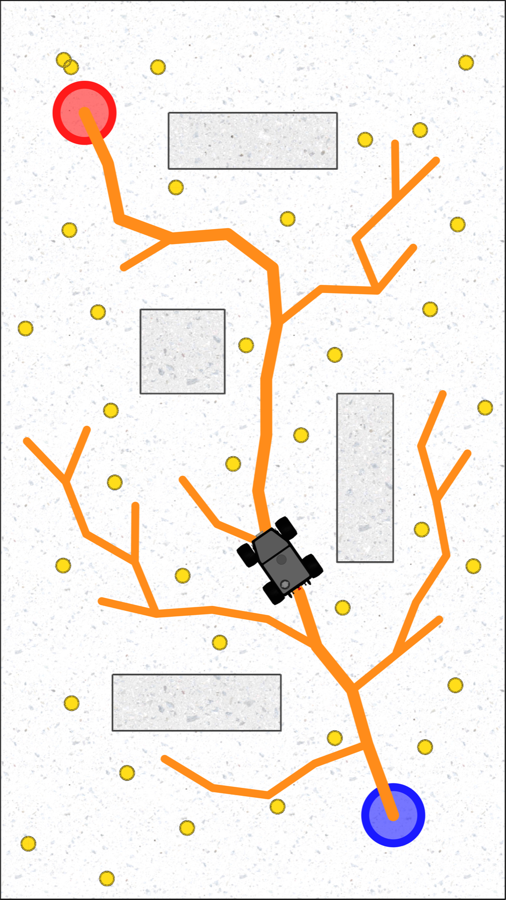}}%
    \hspace{0.05\columnwidth}%
    \raisebox{-0.5\height}{%
        \includegraphics[width=0.26\columnwidth]
            {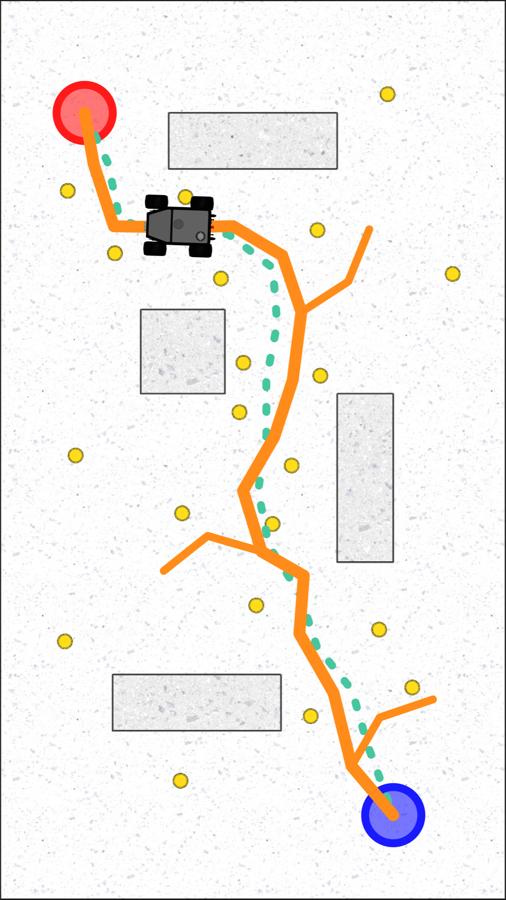}}%
    \hspace{0.02\columnwidth}%
    \raisebox{-0.5\height}{%
        \includegraphics[width=0.24\columnwidth]
            {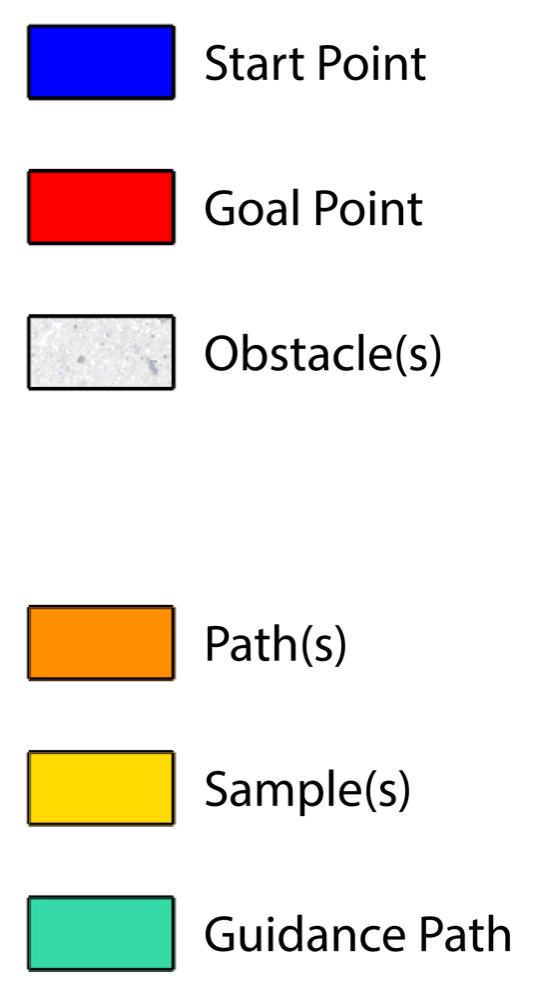}}
    \caption{
        Motion planning for a wheeled robot without (left) and with (right) path-guided sampling.
        Uniform sampling leads to broad exploration, whereas prioritizing unreached waypoints along a mass-point guide path concentrates the search along this route to the goal and reduces planning time (\hyperref[sec:e3-path-guided-sampling]{Experiment E3}).
    }
    \label{fig:wheeled-robot-path-guided-vs-not}
\end{figure}
With a uniform sampling strategy, many samples may be unhelpful in scenes with narrow passages or may lead the exploration into areas that do not necessarily lead to the goal configuration.
Using the idea of RRT-Path~\cite{vonasek2009rrtpath}, we implement a path-guided sampling strategy that turns the planner into a two-level planner that first generates a guide path and then uses this guide to focus sampling on promising regions when planning for the actual robot.

The first stage of the two-level planner uses a simple mass-point planner to obtain an initial path $P_{\mathrm{init}}$ from the start state to the goal state.
The mass-point model approximates the robot's shape and its ability to climb steps or traverse obstacles.
We model this ability through a jump whose maximum height is configured to reflect the robot's capabilities.
The mass-point model jumps whenever it detects a step within this height limit.
This jump represents the robot's ability to overcome height differences, rather than necessarily corresponding to a physical jumping motion.

Before using the mass-point path $P_{\mathrm{init}}$ for guided sampling, we refine it through randomized shortcutting \cite{geraerts2007highquality} to reduce unnecessary detours, yielding the refined path $P_{\mathrm{ref}}$:
Each shortcut attempt selects two non-adjacent path nodes and constructs a cubic reference trajectory matching their planar positions and velocities. %
The reference is tracked using inverse-dynamics feedforward and proportional--derivative feedback, with controls clipped to the admissible actuator range.
Accepted shortcuts replace the original path segment; otherwise, the segment is retained.

During second-stage planning, we use the refined path $P_{\mathrm{ref}}$ to guide sampling, as illustrated in \cref{fig:wheeled-robot-path-guided-vs-not}.
We draw a sample from the unreached guide waypoints $q_G$ of $P_{\mathrm{ref}}$ with probability $\rho_{\mathrm{guide}}$ and perturb it with noise with probability $\rho_{\mathrm{noise}}$.
By choosing \(\rho_{\mathrm{guide}}<1\), we retain a nonzero probability of global sampling, allowing the planner to leave an inaccurate guide and discover routes that are infeasible for the mass-point approximation but feasible for the robot and its policy.
As the tree expands, a guide waypoint is removed once the tree contains a node whose distance to that waypoint is below a threshold \(\varepsilon_{\mathrm{guide}}\).
This prevents oversampling areas where the tree already has enough coverage.
If the tree reaches a waypoint farther along the guide before reaching the preceding waypoints, those preceding waypoints are also removed, allowing the planner to exploit shortcuts and avoid sampling from bypassed segments.

When path-guided sampling is active but the guide has no more waypoints to sample from, we enter the random-sampling branch.
Within this branch, the goal is sampled with conditional probability \(\rho_{\mathrm{goal}}\) \cite{lavalle2006planning}; if the goal is not selected, the target is sampled uniformly from \(\mathcal{Z}\):
\begin{equation}
\begin{aligned}
    P(\text{guide}) &= \rho_{\mathrm{guide}}, \\
    P(\text{goal}) &= (1-\rho_{\mathrm{guide}})\rho_{\mathrm{goal}}, \\
    P(\text{uniform}) &= (1-\rho_{\mathrm{guide}})(1-\rho_{\mathrm{goal}}).
\end{aligned}
\label{eq:path-guided-sampling-probabilities}
\end{equation}

\subsection{Steering}
\label{sec:controllers}
The control architecture we apply for the planning is split into high- and low-level controllers:
The high-level controllers generate commands $c_t$ that the low-level controllers then convert into actuator inputs $a_t$ to produce the desired motion and move towards the sampled state $z_\mathrm{rnd}$.
The controllers in our framework are modular; ideally, the resulting robot action brings us closer to the sampled state.

\paragraph*{Guided Monte Carlo}
\label{par:guided-monte-carlo-as-a-high-level-controller}
The Guided Monte Carlo (GMC) controller samples \(k\) command–duration pairs and propagates each command through the system for the corresponding duration, and returns the candidate where the resulting state has the smallest distance to the target \cite{lavalle2001kinodynamic,honig2024motionplanning}.
GMC does not make any assumptions about the robot’s dynamics or the control policy, and can thus be applied to any robot–policy combination, and can operate at either level:
As a high-level controller, the commands are, e.g., yaw rate and speed, and are passed to a low-level controller.
As a low-level controller, its sampled commands are directly actuator inputs.
This flexibility makes it a versatile fallback when no robot- or policy-specific controller is available.
However, GMC is computationally inefficient: randomly sampled commands often make little progress toward the target configuration $z_t$.
\subsubsection{High-Level Controllers}
In addition to GMC as a high-level controller, we elaborate on two more targeted high-level controllers that we use in this work in \cref{subsec:controller_details}.

\subsubsection{Low-Level Controllers}
\label{sec:low-level-controllers}

In addition to GMC as a low-level controller, we use pretrained policies as low-level controllers. Each policy maps environment observations and a high-level command (usually specifying the desired forward speed and yaw rate) to actuator-level inputs that drive the robot accordingly.
The observations may need to be adapted per policy, but the planning algorithm remains unchanged, provided that the policy accepts the required command representation and produces suitable actuator-level control inputs.

\subsection{Distance Metric}
\label{sec:distance-metric}
For nearest-node selection and goal evaluation, we use a weighted squared Euclidean distance between reduced planning states $z_i=(p_i,v_i,\psi_i)$ and $z_j=(p_j,v_j,\psi_j)$,
\begin{equation}
    d(z_i,z_j)
    =
    w_p \left\lVert \Delta p_{ij} \right\rVert_W^2
    +
    w_v \left\lVert \Delta v_{ij} \right\rVert_2^2
    +
    w_\psi \Delta\psi_{ij}^{2},
    \label{eq:weighted-squared-distance}
\end{equation}
where $\Delta p_{ij} = p_i - p_j$, $\left\lVert \cdot \right\rVert_W^2$ denotes the squared norm weighted by the diagonal matrix $W = \mathrm{diag}(w_x, w_y, w_z)$, and the heading difference is wrapped as
\begin{equation}
    \Delta\psi_{ij}
    =
    \min
    \left(
        \left|\psi_i-\psi_j\right|,
        2\pi-\left|\psi_i-\psi_j\right|
    \right).
\label{eq:wrapped-heading-difference}
\end{equation}
This metric is a cheap proxy for connectivity rather than a measure of true reachability, which is instead assessed by the simulator.
The weights $w_p, w_v, w_\psi$ determine the relative importance of position, velocity, and heading.
Nearest-neighbor selection is sensitive to these weights: an unsuitable weighting may select a spatially close node that has an incompatible velocity or orientation, or may overemphasize one component of the planning state.
We additionally consider a height-sensitive variation in which \(w_z \gg w_x, w_y\).
This penalizes vertical displacement more strongly than horizontal displacement.
It can prevent nodes on different terrain levels from being treated as close merely because their horizontal positions are similar.
However, excessive vertical weighting may also discourage exploration of transitions involving ramps, stairs, jumps, or climbing.
The distance weights must thus be selected with respect to the robot--policy capabilities.

\subsection{Pruning the Search Tree}
\label{sec:pruning-the-search-tree}
During our experiments, we observed that the planner can spend excessive time exploring tree branches which lead to dead ends, such as a robot falling or tripping. To alleviate this problem, we implemented a pruning mechanism, inspired by RC-RRT’s \cite{cheng2002resolution} constraint-violation-frequency (CVF) scheme, which propagates failure penalties to ancestors and probabilistically deprioritizes high-failure nodes. In our implementation, we do not deprioritize nodes, but instead prune them from the tree entirely. We prune only leaf nodes. However, if a node becomes childless due to pruning, it is also eligible for pruning. Thus, the pruning process can propagate up the tree.

\section{Experiments and Results}
\label{sec:experiments-and-results}

All experiments were conducted on workstations with an AMD Ryzen 7 5800X CPU, 32\,GiB of RAM, and an NVIDIA GeForce RTX 3080 Ti GPU.
We use MuJoCo 3.10.0 \cite{todorov2012mujoco} as the simulator.
For the experiments, we support three task types: one motion-planning task, $\operatorname{MoveTo}$, and two manipulation tasks, $\operatorname{PickUp}$ and $\operatorname{Place}$. By default, we use \emph{Aicardi}~\cite{aicardi1995steering} as the high-level controller, paired with a control policy as the low-level controller. If no policy is available, we use GMC as the fallback low-level controller. We use a planning budget of $30{,}000$ RRT iterations per motion planning task.

Throughout this section, \emph{Success} is the percentage of tasks completed within the planning budget; \emph{Completed} reports successful/attempted runs.
\emph{Time} and \emph{Preplan} are mean wall times for planning and mass-point preplanning, respectively, and \emph{$q_{25}$} and \emph{$q_{75}$} are planning-time percentiles.

\subsection{Controller Details}
\label{subsec:controller_details}
As low-level controllers, we use the pretrained policy of Yoon et al.~\cite{yoon2026teachingrobotslikedogs} for the quadruped (\texttt{go2\_parkour}) and the perceptive locomotion policy of Zhu et al.~\cite{zhu2026hiking} for the humanoid Unitree G1 (\texttt{g1\_hiking})\footnote{We use the checkpoint from Project Instinct at \url{https://github.com/project-instinct/InstinctMJ}.}.

In addition to the Guided Monte Carlo steering as a high-level controller, we compare two additional controllers that are more specific to the quadruped and humanoid:

\paragraph*{Heading-Error Controller}
\label{par:heading-error-controller}
This controller uses a simple heuristic based on the robot's heading error and planar distance to the target. It initially prioritizes rotation toward the target and progressively increases the forward velocity as the heading error decreases. This behavior prevents the robot from moving rapidly in the wrong direction while still enabling efficient forward motion once it is sufficiently aligned with the target.

\paragraph*{Aicardi Controller}
\label{par:aicardi-controller}

The unicycle controller of Aicardi et al. \cite{aicardi1995steering} generates a forward velocity $u$ and yaw rate $\omega$:
\begin{equation}
\begin{aligned}
    u &= \gamma e \cos\alpha, \\
    \omega &= k\alpha + \gamma \frac{\cos\alpha\sin\alpha}{\alpha}
      (\alpha + h\theta),
\end{aligned}
\end{equation}
where $e$ is the planar distance to the target, $\alpha$ is the heading error relative to the target direction, and $\theta$ is the approach-orientation error, with angular conventions following \cite{aicardi1995steering}.

\subsection*{E1: Environment-Modification-Conditioned Reachability}
\label{sec:e1-environment-modification-conditioned-reachability}

\begin{figure}[t]
    \centering
    \includegraphics[width=\columnwidth]
        {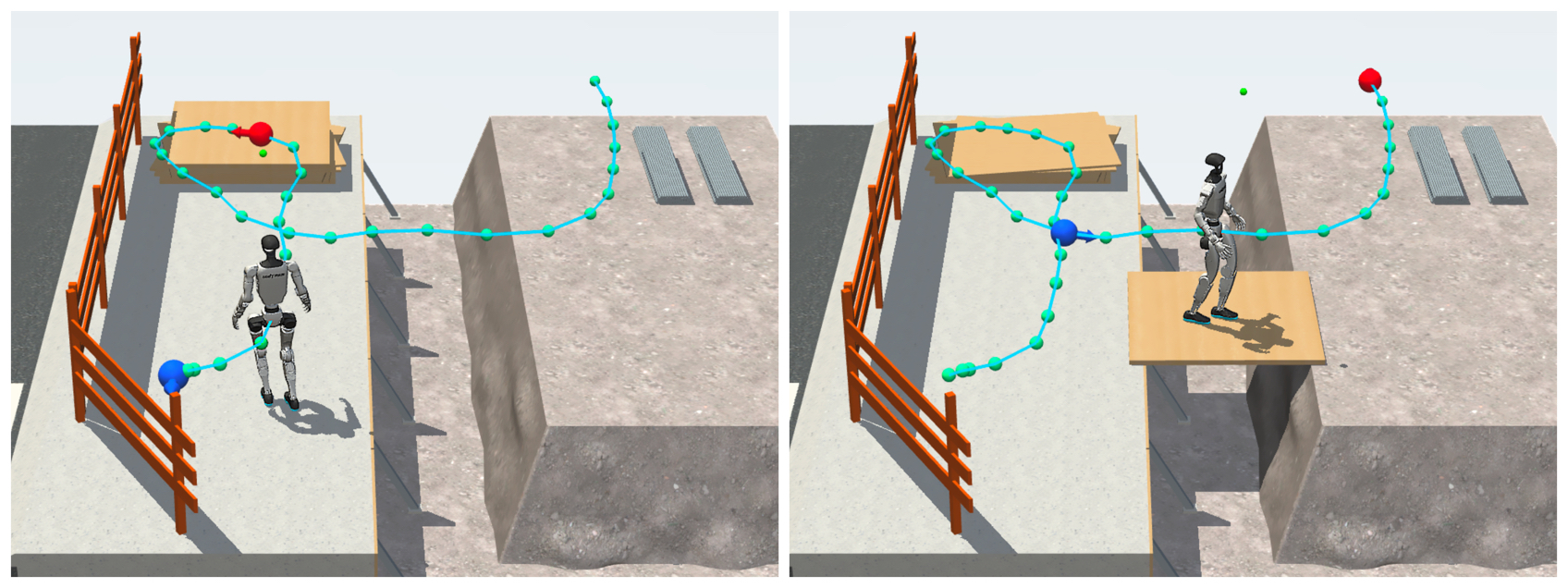}
    \caption{
    A humanoid in the construction-site scene, where it must first pick up and place a plank over the gap to reach the goal location.
    The blue line indicates the planned path.
    The dark blue dot indicates the start state, and the red dot indicates the goal state of a task.
    In this example, we have the task plan:
    1) $\operatorname{MoveTo}$ the plank,
    2) $\operatorname{PickUp}$ the plank,
    3) $\operatorname{MoveTo}$ the gap,
    4) $\operatorname{Place}$ the plank over the gap, and
    5) $\operatorname{MoveTo}$ the goal location.
    }
    \label{fig:humanoid-build-bridge}
\end{figure}

This experiment demonstrates tasks in which reachability depends on modifying the environment. \Cref{fig:humanoid-build-bridge} shows an example of such a task and how it enables the robot to reach a new area. We consider three scenes requiring sequences of navigation and manipulation actions.
We use two bridge-building scenes in which a single plank must be placed across a gap; the more challenging scene requires placing it at an elevated location.
In the scene \emph{Stair}, the robot must place three blocks sequentially to construct a staircase to an elevated platform.
We evaluate \texttt{car}, \texttt{go2\_parkour}, and \texttt{g1\_hiking} over five runs per task plan, excluding \texttt{car} from the \emph{Stair} scene.
Without environment modifications, no evaluated robot reached the goal within the planning budget of $30{,}000$ RRT iterations in any run.
With the prescribed modifications, every evaluated robot--scene combination succeeded in all five runs.

\subsection*{E2: Policy-Conditioned Reachability}
\label{sec:e2-policy-conditioned-reachability}
\begin{figure}[t]
    \centering
    \includegraphics[width=\columnwidth]
        {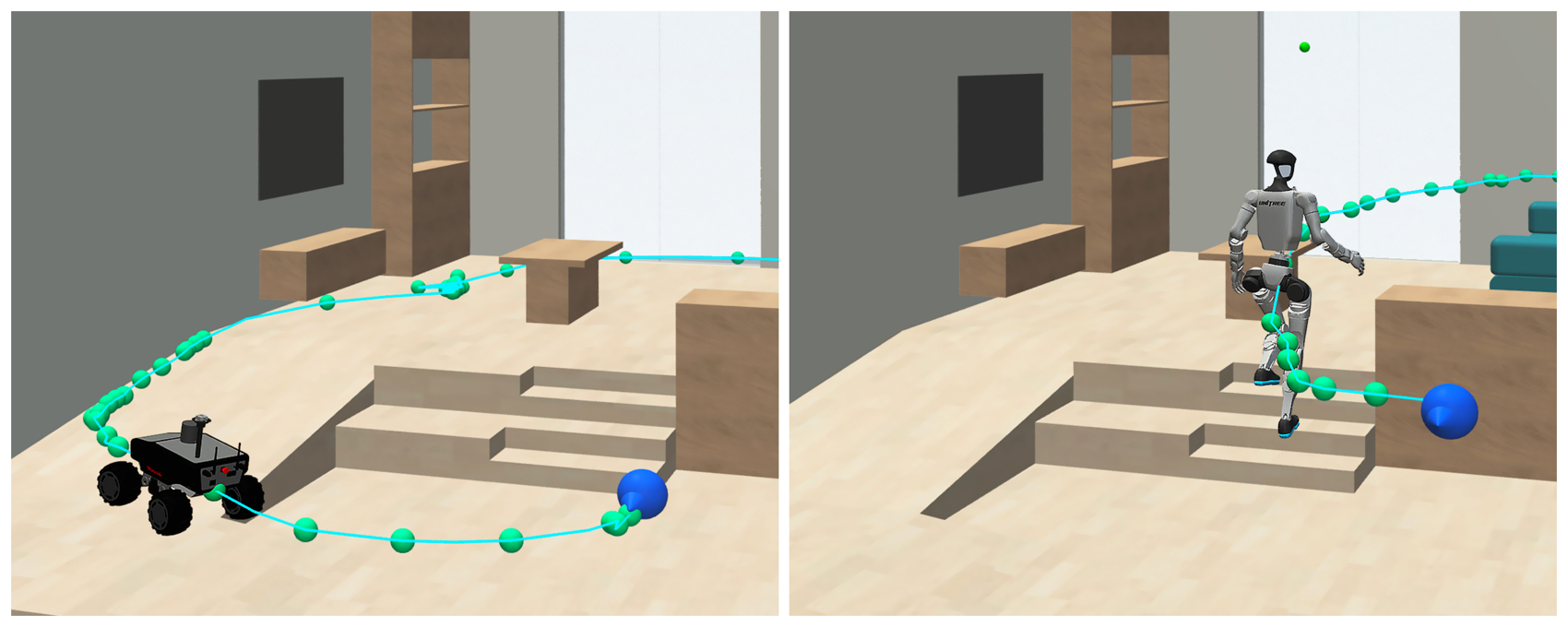}
    \caption{In this scene, the wheeled robot must take the longer route via the ramp to reach the elevated platform, whereas the humanoid can follow a shorter route by climbing the stairs.}
    \label{fig:different-capabilities}
\end{figure}

We evaluate how robot--policy capabilities affect reachability across four scenes: \textit{flat}, \textit{ramp}, \textit{low stair}, and \textit{high stair}.
The obstacle-free flat scene serves as a baseline, while the remaining scenes test terrain-traversal capabilities.
All robot--policy combinations achieved 100\% success on flat terrain.
On the ramp and low stair, \texttt{go2\_parkour} and \texttt{g1\_hiking} achieved 100\% success, while \texttt{car} achieved 90\% and 70\%, respectively, and \texttt{go2} failed in every trial.
On the high stair, only \texttt{g1\_hiking} succeeded; all other combinations failed.
These results show that reaching a goal depends on the robot--policy combination, and qualitatively show in \cref{fig:different-capabilities} that our approach can exploit each policy's capabilities by taking paths that are appropriate for the robot's capabilities.

\subsection*{E3: Path-Guided Sampling}
\label{sec:e3-path-guided-sampling}

We compare path-guided and unguided sampling and evaluate which variation of path-guided sampling provides the best performance. For the path-guided configurations we used $\rho_{guide}=0.5$ and $\rho_{noise}=0.2$.
We also assess the planner’s robustness to misleading guide paths.
The evaluation includes four path-guided sampling configurations.
The default \emph{path-guided} configuration, described in \cref{par:path-guided-sampling}, samples from the next few unreached guide waypoints.
The \emph{whole-path} configuration samples from all remaining waypoints, while \emph{no-shortcut} retains bypassed waypoints instead of discarding them as the tree progresses.
The \emph{misleading path} configuration deliberately uses a guide path that does not lead to the goal.
Across 660 motion-planning problems per configuration and all robots, %
path-guided sampling increases success and reduces mean planning time compared with unguided sampling (\cref{tab:e3-configuration-comparison}).
Including \(6.3\,\mathrm{s}\) of mean preplanning time, the reduction in mean total planning time is $\approx63.4\%$. %

We next compare the three path-guided sampling variants: \emph{path-guided}, \emph{whole-path}, and \emph{no-shortcut}. All three achieve similar success rates, but the default \emph{path-guided} configuration yields the lowest mean and 75th-percentile planning times, providing the best overall trade-off.
As shown in \cref{tab:e3-configuration-comparison}, a \emph{misleading path} reduces the success rate and increases the mean planning time.
Nevertheless, the planner still finds a solution in most trials.
This demonstrates that misleading guidance does not generally prevent the planner from recovering, although recovery comes at a substantially increased computational cost.

\begin{table}[t]
    \centering
    \caption{
        Comparison of path-guided sampling configurations, aggregated across scenes and robots.
    }
    \label{tab:e3-configuration-comparison}
    \sisetup{detect-weight=true}
    \begin{tabular*}{\columnwidth}{@{\extracolsep{\fill}}l
        S[table-format=2.1, table-space-text-post={\%}]
        S[table-format=3.1]
        S[table-format=2.1]
        S[table-format=3.1]
        S[table-format=1.1]@{}}
        \toprule
        \textbf{Config} &
        {\textbf{Success}} &
        {\textbf{Time [s]}} &
        {\textbf{$q_{25}$ [s]}} &
        {\textbf{$q_{75}$ [s]}} &
        {\textbf{Preplan [s]}} \\
        \midrule
        Path-Guided
            & \bfseries 99.7{\%}
            & \bfseries 27.5
            & 2.9
            & \bfseries 16.7
            & 6.3 \\
        \addlinespace
        Whole-Path
            & 98.6{\%}
            & 31.0
            & \bfseries 2.5
            & 18.9
            & 6.4 \\
        No-Shortcut
            & 98.2{\%}
            & 30.3
            & 3.7
            & 20.6
            & 6.2 \\
        \addlinespace
        Unguided
            & 95.6{\%}
            & 92.4
            & 7.5
            & 92.0
            & {--} \\
        \addlinespace
        Misleading
            & 92.6{\%}
            & 129.8
            & 11.0
            & 94.9
            & 6.1 \\
        \bottomrule
    \end{tabular*}
\end{table}

\subsection*{E4: General Planning Ability}
\label{sec:e4-general-planning-ability}

Experiment E4 evaluates planning across obstacle-rich and finite-support terrains, elevation changes and mazes.
Using tuned robot-specific configurations, we executed a total of 830 individual motion planning problems across different scenes and robots.
\Cref{tab:e4-general-planning-ability} shows perfect task plan completion across scene types.
Results for representative scenes in \cref{tab:e4-scene-robot-comparison} further show that planning time depends strongly on the scene and robot-policy combination.

\begin{table}[t]
    \centering
    \caption{
        Motion-planning task completion by scene type, aggregated across robots.
    }
    \label{tab:e4-general-planning-ability}
    \begin{tabular*}{\columnwidth}{@{\extracolsep{\fill}}lr@{}}
        \toprule
        \textbf{Scene type} &
        \textbf{Completed} \\
        \midrule
        Obstacle navigation
            & 280/280 \\
        Elevation changes
            & 310/310 \\
        Mazes
            & 120/120 \\
        \addlinespace
        Planar baseline
            & 120/120 \\
        \midrule
        \textbf{Overall}
            & \textbf{830/830} \\
        \bottomrule
    \end{tabular*}
\end{table}

\begin{table}[t]
    \centering
    \caption{
        Task plan completion and planning times for representative scenes.
        Wall times are summed over motion-planning tasks within each successful task plan.
        A task plan is successful only if all its motion-planning tasks succeed and no error occurs.
        The construction-site task plan contains three motion-planning tasks; the other listed task plans contain one.
        The construction-site and living-room scenes are illustrated in \cref{fig:humanoid-build-bridge,fig:e4-planned-path-replay}, respectively.
    }
    \label{tab:e4-scene-robot-comparison}
    \begin{tabular*}{\columnwidth}{@{\extracolsep{\fill}}llr
        S[table-format=3.1]@{\extracolsep{0pt}${}\pm{}$}S[table-format=3.1]@{}}
        \toprule
        \textbf{Scene} &
        \textbf{Robot} &
        \textbf{Completed} &
        \multicolumn{2}{c@{}}{\textbf{Time [s]}} \\
        \midrule
        construction site
            & g1\_hiking   & 10/10 & 41.1  & 57.7  \\
            & go2\_parkour & 10/10 & 132.4 & 323.2 \\
        \addlinespace
        living room
            & g1\_hiking   & 10/10 & 11.3  & 18.9  \\
            & go2\_parkour & 10/10 & 163.0 & 150.3 \\
        \addlinespace
        small spiral ramp
            & car          & 10/10 & 25.5 & 21.8  \\
            & g1\_hiking   & 10/10 & 92.9 & 110.3 \\
            & go2\_parkour & 10/10 & 23.0 & 19.3  \\
        \addlinespace
        labyrinth
            & car          & 10/10 & 10.2 & 6.1  \\
            & g1\_hiking   & 10/10 & 2.6  & 0.4  \\
            & go2\_parkour & 10/10 & 55.4 & 43.3 \\
        \addlinespace
        flat (baseline)
            & car          & 20/20 & 5.3 & 3.0 \\
            & g1\_hiking   & 20/20 & 2.1 & 0.8 \\
            & go2\_parkour & 20/20 & 6.4 & 2.3 \\
        \bottomrule
    \end{tabular*}
\end{table}

\begin{figure}[t]
    \centering
    \includegraphics[width=\columnwidth]{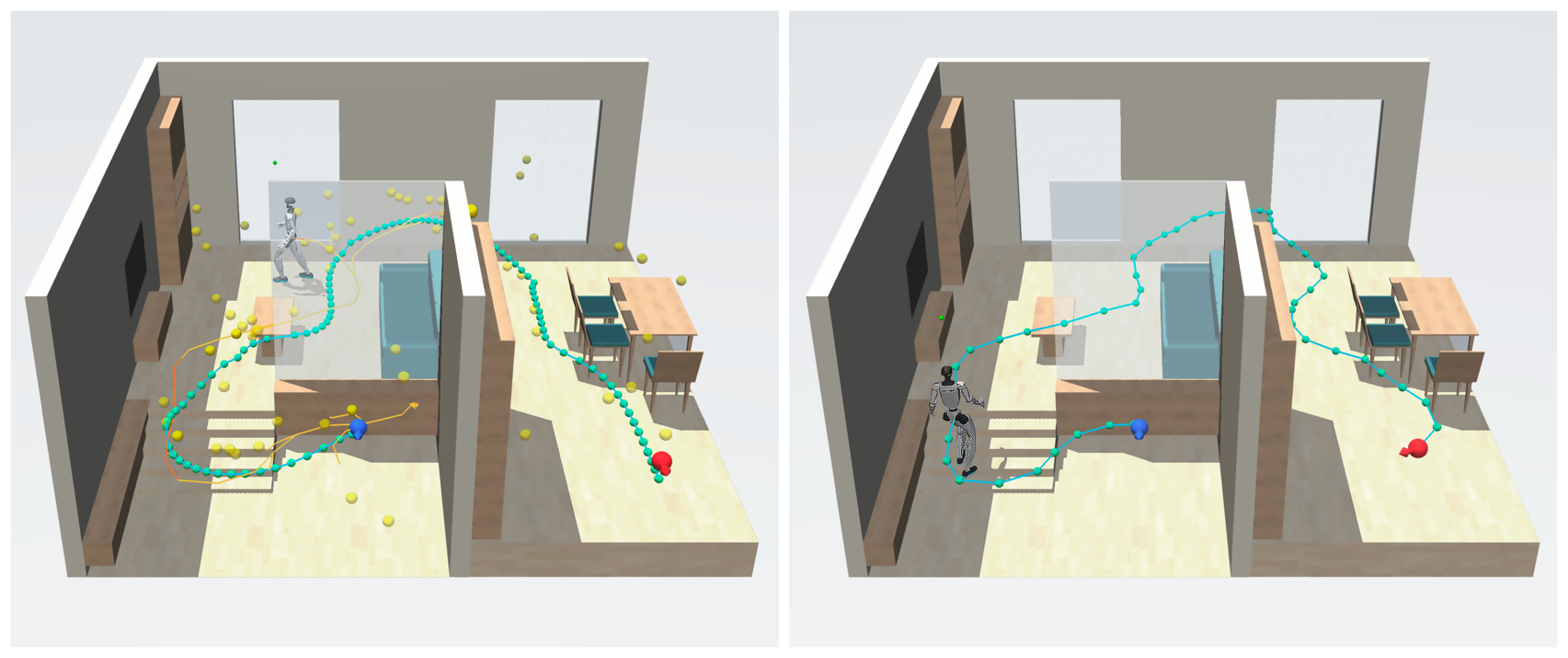}
    \caption{
    This figure shows a humanoid in the living-room scene.
    The left image shows the planning process:
    The blue path is the guidance path generated using a simplified mass-point model,
    yellow dots indicate the sampled configurations in 3D space, which are used to grow the RRT.
    The thin yellow lines indicate the edges of the RRT that is built.
    The right image shows the humanoid executing the resulting path by walking from the start configuration (blue dot) to the goal configuration (red dot).
    }
    \label{fig:e4-planned-path-replay}
\end{figure}

\subsection*{E5: Steering Methods}
\label{sec:e5-steering-methods}
Experiment E5 compares the Guided Monte Carlo, Aicardi, and Heading-Error high-level controllers across scenes. %
\Cref{tab:e5-robot-configuration-comparison} shows that Aicardi achieves the lowest planning wall time across both robots while maintaining a success rate comparable to that of Guided Monte Carlo, with only a slight reduction for \texttt{g1\_hiking}. It therefore provides the best overall efficiency–reliability trade-off.

\begin{table}[t]
    \centering
    \caption{
        Comparison of high-level controllers by robot, aggregated across scenes. Times are reported as mean $\pm$ standard deviation.
        }
    \label{tab:e5-robot-configuration-comparison}
    \begin{tabular*}{\columnwidth}{@{\extracolsep{\fill}}llr
        S[table-format=3.1]@{\extracolsep{0pt}${}\pm{}$}S[table-format=3.1]@{}}
        \toprule
        \textbf{Robot} &
        \textbf{Configuration} &
        \textbf{Completed} &
        \multicolumn{2}{c@{}}{\textbf{Time [s]}} \\
        \midrule
        go2\_parkour
            & Aicardi
            & \textbf{205/205}
            & \bfseries 24.9 & \bfseries 49.1 \\
            & Heading-Error
            & 204/205
            & 27.6 & 136.6 \\
            & \shortstack[l]{Guided Monte Carlo}
            & \textbf{205/205}
            & 83.7 & 292.9 \\
        \midrule
        g1\_hiking
            & Aicardi
            & 201/205
            & \bfseries 34.7 & \bfseries 160.1 \\
            & Heading-Error
            & 199/205
            & 36.7 & 180.9 \\
            & \shortstack[l]{Guided Monte Carlo}
            & \textbf{204/205}
            & 105.9 & 471.1 \\
        \bottomrule
    \end{tabular*}
\end{table}

\subsection*{E6: Patch Sampling}
Experiment E6 compares sampling in 3D space with sampling on predefined traversable patches on 660 individual motion planning tasks per configuration.
In patch sampling, reduced states are sampled with probability proportional to the surface area of the respective patch.
\Cref{tab:e6-patch-sampling} shows that sampling in 3D performs better, with a higher success rate and a lower mean planning time. The difference is most pronounced in the mean preplanning time.%

The difference in performance between the two approaches is biggest in the \textit{spiral ramp} scene (\cref{tab:spiral-ramp-sampling-comparison}).
We attribute this behavior to the sampling distribution and nearest-neighbor selection that differ significantly between the approaches.
While 3D sampling draws samples uniformly from the scene’s bounding box, patch-based sampling selects patches in proportion to their surface area, favoring the large floor beneath the spiral ramp over the smaller elevated platforms.
Additionally, samples on the elevated patches often select nearest neighbors on the traversable surface below, triggering expansion from that lower level instead of the elevated patches.
Upward growth therefore requires samples sufficiently close to the ascending branch, which are rare because elevated patches receive fewer samples.
This prolongs preplanning across all robots: the mass-point planner takes longer to find a route to the elevated platform and generate the guide path for subsequent robot planning.
Without path-guided sampling, this effect would likely be more pronounced, leading to substantially longer mean planning times for the quadruped, humanoid, and wheeled robots.

\begin{table}[t]
    \centering
    \caption{
        Comparison of 3D and patch-based sampling, aggregated across scenes and robots.
        Total is the sum of the mean planning time and mean preplan time.
    }
    \label{tab:e6-patch-sampling}
	\resizebox{\columnwidth}{!}{%
		\begin{tabular}{@{}l
			S[table-format=2.1{\%}]
			S[table-format=2.1]
			S[table-format=1.1]
			S[table-format=2.1]
			S[table-format=2.1]
			S[table-format=2.1]@{}}
			\toprule
			\textbf{Config} &
			{\textbf{Success}} &
			{\textbf{Time [s]}} &
			{\textbf{$q_{25}$ [s]}} &
			{\textbf{$q_{75}$ [s]}} &
			{\textbf{Preplan [s]}} &
			{\textbf{Total [s]}} \\
			\midrule
			3D
				& \bfseries 99.7{\%}
				& \bfseries 27.5
				& \bfseries 2.9
				& \bfseries 16.7
				& \bfseries 6.3
				& \bfseries 33.8 \\
			Patch
				& 99.1{\%}
				& 35.1
				& 3.8
				& 20.3
				& 24.4
				& 59.5 \\
			\bottomrule
		\end{tabular}%
	}
\end{table}

\begin{table}[t]
	\centering
	\small
	\setlength{\tabcolsep}{4pt}
	\caption{
        Comparison of 3D and patch-based sampling in the \textit{spiral ramp} scene, with 10 runs per robot and sampling method.
        Total is the sum of the mean planning time and mean preplan time.
    }
	\label{tab:spiral-ramp-sampling-comparison}
    \resizebox{\columnwidth}{!}{%
        \begin{tabular}{@{}ll
            S[table-format=3.0{\%}]
            S[table-format=3.1]
            S[table-format=2.1]
            S[table-format=3.1]@{}}
            \toprule
            \textbf{Robot}	& \textbf{Config}	& {\textbf{Success}}	& {\textbf{Time [s]}}	& {\textbf{Preplan [s]}}	& {\textbf{Total [s]}} \\
            \midrule
            \multirow{2}{*}{car}
            & 3D	& 100{\%}	& \bfseries 21.1	& \bfseries 23.9	& \bfseries 45.0 \\
            & Patch	& 100{\%}	& 36.9				& 70.9				& 107.8 \\
            \midrule
            \multirow{2}{*}{go2\_parkour}
            & 3D	& 100{\%}	& 63.1				& \bfseries 15.2	& \bfseries 78.3 \\
            & Patch	& 100{\%}	& \bfseries 59.7	& 69.5				& 129.2 \\
            \midrule
            \multirow{2}{*}{g1\_hiking}
            & 3D	& 100{\%}	& \bfseries 111.9	& \bfseries 16.5	& \bfseries 128.4 \\
            & Patch	& 100{\%}	& 226.9				& 79.2				& 306.1 \\
            \bottomrule
        \end{tabular}%
    }
\end{table}

\section{Discussion \& Limitations}
\label{sec:limitations}
In this work, we introduced a planning approach that we see as an initial step towards a task-and-motion planning approach, by enabling planners to plan in the simulated environment and also modify it.
We do currently still assume a predefined, fixed task plan.
We focus here on the motion-planning aspect: we do not yet generate, select, or reorder tasks during planning.
Nevertheless, the proposed approach should also provide a basis to work toward a full task-and-motion planning approach.

We show in extensive experiments that the planning performance is adequate in many settings. %
Given recent trends, the planner should in the future leverage modern parallelism more, both from CPUs and GPUs.
This would allow us to decrease the planning time significantly through more efficient rollouts, faster simulation, and task-level parallelism.
Moreover, planner performance depends on robot-specific parameter tuning, and no formal completeness result is established in this work: best-input extension with a fixed step is known not to be probabilistically complete~\cite{kunz2014incomplete}, and the completeness of the best-of-$k$ random command--duration propagation we use is an open question~\cite{kleinbort2019completeness}.
Finally, the path-guided sampling assumes that the mass-point path can be transferred to other robots.
This assumption may not hold in all cases.

Each robot currently uses a single policy.
In future work, the planner could choose from a discrete skill index in the planning process directly among several policies per robot, as the learned navigation policy of ANYmal Parkour does for a fixed skill catalog~\cite{hoeller2024parkour}, with the same simulator-in-the-loop validation and without retraining.
Similarly, while we plan using policies that have been validated in the physical world, we only validated feasibility in the physics simulator, not on a physical system in the real world.
We assume that the goal-conditioned policies transfer to the real systems, but we did not test this in this work.

\section{Conclusion}
We presented a kinodynamic RRT planner that uses physics simulation to exploit robot and control-policy capabilities as well as environmental changes.
The planner evaluates motions through simulated rollouts and carries manipulation outcomes into subsequent planning.
Experiments with humanoids, quadrupeds, and wheeled robots demonstrate reliable planning across diverse 3D environments.
Guiding sampling with paths from a simplified mass-point model reduced mean planning time and increased the success rate.
We see this work as a promising foundation for future work in policy-in-the-loop task-and-motion planning, and we plan on improving computational efficiency and validating execution on physical robots.

\section{Acknowledgments}
\label{sec:acknowledgments}
We thank Taerim Yoon and Sitong Chen for providing pretrained control policies, as well as the team at the Computational Robotics Lab (CRL) for their support.
Yijiang Huang is supported by the SNSF Ambizione Grant (Grant No. 223384).\\
The authors used AI tools to assist with coding, data processing, and scene creation, excluding visualization scenes.

\bibliographystyle{IEEEtran}
\bibliography{IEEEabrv,references}

\end{document}